\documentclass[a4paper, 10pt, conference]{ieeeconf}      
\usepackage{FG2021}

\FGfinalcopy 

\IEEEoverridecommandlockouts                              
\usepackage[utf8]{inputenc} 
\usepackage[T1]{fontenc}    
\usepackage{hyperref}       
\usepackage{url}            
\usepackage{booktabs}       
\usepackage{amsfonts}       
\usepackage{nicefrac}       
\usepackage{microtype}      
\usepackage{epsfig}
\usepackage{graphicx}
\usepackage{amsmath}
\usepackage{amssymb}
\usepackage{float}
\usepackage{color}
\usepackage{colortbl}
\usepackage{breqn}
\usepackage{rotating}
\usepackage{times}
\usepackage{epsfig}
\let\labelindent\relax      
\usepackage{enumitem}
\usepackage{makecell}
\usepackage{lineno}
\usepackage{tabularx}
\usepackage{multirow}
\usepackage[maxnames=100]{biblatex}
\usepackage{subfigure}
\usepackage{wrapfig}
\usepackage{comment}
\usepackage{bbm}
\usepackage{bm}

\def\0{{\bf 0}}
\def\1{{\bf 1}}

\newcommand{\etal}[1]{\textit{et al.}}

\definecolor{red}{rgb}{0.95,0.4,0.4}
\definecolor{purered}{rgb}{1,0,0}
\definecolor{blue}{rgb}{0.4,0.4,0.95}
\definecolor{darkblue}{rgb}{0,0,0.8}
\definecolor{darkred}{rgb}{1,0,0}
\definecolor{darkgreen}{rgb}{0,0.5,0}
\definecolor{grey}{rgb}{0.6,0.6,0.6}
\definecolor{col1}{RGB}{232, 161, 148}
\definecolor{col2}{RGB}{148, 187, 232}
\definecolor{lightgrey}{rgb}{0.85,0.85,0.85}
\definecolor{lightlightgrey}{rgb}{0.9,0.9,0.9}
\definecolor{verylightBG}{rgb}{0.9,0.99,0.99}
\definecolor{darkgreen}{rgb}{0.3, 0.75, 0.3}
\definecolor{darkgrey}{rgb}{0.8,0.35,0.35}
\DeclareUnicodeCharacter{0301}{\'{e}}
\graphicspath{{./figures/}}   %
\AtBeginBibliography{\small}
\def\FGPaperID{205} 

\title{\LARGE \bf
A Synthesis-Based Approach for Thermal-to-Visible Face Verification
}

\author{\parbox{16cm}{\centering
    {\large Neehar Peri$^{1,2}$, Joshua Gleason$^{1,2}$, Carlos D. Castillo$^{1,3}$, Thirimachos Bourlai$^4$, \\
    Vishal M. Patel$^{1,3}$, Rama Chellappa$^{1,3}$}\\
    {\normalsize
    $^1$ MUKH Technologies,
    $^2$ University of Maryland,
    $^3$ Johns Hopkins University,
    $^4$ University of Georgia}}
}

\begin{document}
\IEEEoverridecommandlockouts\pubid{\makebox[\columnwidth]{978-1-6654-3176-7/21/\$31.00~\copyright{}2021 IEEE \hfill} \hspace{\columnsep}\makebox[\columnwidth]{ }}

\ifFGfinal
\thispagestyle{empty}
\pagestyle{empty}
\else
\author{Anonymous FG2021 submission\\ Paper ID \FGPaperID \\}
\pagestyle{plain}
\fi
\maketitle

\begin{abstract}
In recent years, visible-spectrum face verification systems have been shown to match the performance of experienced forensic examiners. However, such systems are ineffective in low-light and nighttime conditions. Thermal face imagery, which captures body heat emissions, effectively augments the visible spectrum, capturing discriminative facial features in scenes with limited illumination. Due to the increased cost and difficulty of obtaining diverse, paired thermal and visible spectrum datasets, not many algorithms and large-scale benchmarks for low-light recognition are available. This paper presents an algorithm that achieves state-of-the-art performance on both the ARL-VTF and TUFTS multi-spectral face datasets. Importantly, we study the impact of face alignment, pixel-level correspondence, and identity classification with label smoothing for multi-spectral face synthesis and verification. We show that our proposed method is widely applicable, robust, and highly effective. In addition, we show that the proposed method significantly outperforms face frontalization methods on profile-to-frontal verification. Finally, we present MILAB-VTF(B), a challenging multi-spectral face dataset that is composed of paired thermal and visible videos. To the best of our knowledge, with face data from 400 subjects, this dataset represents the most extensive collection of indoor and long-range outdoor thermal-visible face imagery. Lastly, we show that our end-to-end thermal-to-visible face verification system provides strong performance on the MILAB-VTF(B) dataset.


\end{abstract}

\section{Introduction}
Face verification is concerned with the task of identifying if two face images correspond to the same identity (one-to-one matching). Web-scale visible-spectrum datasets and advances in training deep convolutional neural networks (DCNNs) have yielded significant improvements in face verification, matching the  performance of forensic experts \cite{phillips2018_forensic}. However, methods trained on visible-band face images often fail to generalize to low-light and nighttime conditions \cite{bourlai16_acrossspectrum, bourlai16_outsidevisible}. Thermal imagery, particularly in the Long-Wave Infra-Red (LWIR, 7$\mu$m - 14$\mu$m) and Mid-Wave Infra-Red (MWIR, 3$\mu$m - 5$\mu$m) \cite{bourlai12_difficultenvironments, poster21_arlvtf} bands, addresses limitations of the visible spectrum in low-light applications. Thermal imagery effectively captures discriminative information from body heat signatures. In order to leverage visible spectrum face verification pipelines in low-light scenes, recent works have proposed the task of thermal-to-visible spectrum face synthesis. 

\textbf{Standard Approach.} The standard thermal-to-visible face synthesis pipeline consists of three stages: (1) detect and crop faces, (2) synthesize a corresponding visible-spectrum face for the thermal-spectrum input, and (3) extract discriminative features using a fixed feature extractor trained on visible-spectrum data. Significant distribution shifts between thermal and visible imagery make domain adaptation challenging. In addition to identity labels, recent methods use additional annotations, including pose \cite{zhao18_pim}, part masks \cite{li19_m2fpa, yin20_dagan}, and attributes \cite{di21_attributeguided, he17_ganvfs, he19_highquality}. However, these methods may not effectively scale across diverse datasets and network architectures. The tasks of identity verification \cite{parkhi15_vggface, deng19_arcface} and re-identification \cite{luo19_bagoftricks, khorramshahi20_selfsupervised, khorramshahi19_dualpath} aim to learn robust features for identity matching. Since re-identification and multi-spectral face datasets typically have fewer identities than visible-band verification datasets, we also look to re-identification methods for inspiration. 

\textbf{Standard Datasets.} Currently available paired thermal-visible datasets consist of face images captured under controlled conditions, in terms of  background, illumination, standoff distance and pose. Unlike visible-spectrum face datasets, which often contain thousands of unique identities, with face images captured under constrained and unconstrained conditions, paired thermal-visible datasets are limited in terms of size and diversity in collection conditions. In order for dual-band (visible-thermal) verification pipelines to learn discriminative features, thermal and visible datasets must contain a sufficiently large number of identities.

\textbf{Contributions.} In this paper, we focus on advancing standard practices in thermal-to-visible face synthesis. We show that general-purpose domain adaptation algorithms can be highly effective, achieving state-of-the-art performance on both ARL-VTF \cite{poster21_arlvtf} and TUFTS \cite{panetta20_tufts} datasets. In addition, we show that our method is effective at frontal-to-profile matching, outperforming face frontalization methods \cite{huang17_tpgan, zhao18_pim, li19_m2fpa, yin20_dagan, di21_dalgan}. Lastly, we introduce the largest to-date, long-range, unconstrained paired thermal-to-visible face dataset (MILAB-VTF(B)). The proposed method yields competitive performance on this dataset. We believe this novel dataset will be valuable in closing the data gap between visible-spectrum and multi-spectral datasets. 

\section{Related Works}
In this section, we briefly discuss a limited subset of related work in developing robust end-to-end systems for thermal-to-visible face verification. In particular, we summarize prior face synthesis methods and highlight key features of this large-scale datasets. 

\textbf{Domain Adaptation.}
Pix2Pix \cite{isola17_pix2pix} introduced conditional adversarial networks that learn to adapt by generating realistic samples that match the target distribution. However, Pix2Pix requires paired examples. CycleGAN\cite{zhu17_cyclegan} introduced the cycle-consistency loss to learn a bi-directional mapping between two domains without paired examples. Several methods have adapted the CycleGAN for task-specific applications. CyCADA \cite{hoffman18_cycada} extended CycleGAN to enforce semantic consistency using an auxiliary task loss. Furthermore, BicycleGAN \cite{zhu17_bicyclegan} addressed the problem of mapping two domains to sample diverse examples from a target distribution. More recently, contrastive unpaired translation (CUT) \cite{taesung20_cut} used contrastive learning to maintain the content of the input domain while learning the appearance of the target domain.

\textbf{Face Synthesis.}
 GAN-VFS \cite{he17_ganvfs} proposed an encoder-decoder structure that directly translates polarimetric images to the visible domain while enforcing a perceptual loss on intermediate features so that they closely resemble the intermediate feature embedding from a fined-tuned VGG-16 feature extractor. Similarly, \cite{di19_selfattn} extended CycleGAN to enforce an ID loss to ensure features from synthesized images are close to the corresponding real image features and proposed a feature fusion of both polarimetric and visible image features to improve verification robustness. Unlike most synthesis based-methods, \cite{fondje20_crossdomain} directly adapted intermediate features from both thermal and visible images using truncated fixed feature extractors to learn a domain invariant representation for cross-domain matching. More recently, \cite{di21_attributeguided} introduced a method that incorporates facial attributes by pooling latent features with attribute features and synthesized visible domain images at multiple scales to guide face synthesis effectively. Similarly, \cite{mallat19_cascade} exploited multi-scale information for higher resolution generation with less training data using a series of cascade refinement networks. 
 
 Several methods have investigated the sub-problem of face frontalization for profile-to-frontal matching. TP-GAN \cite{huang17_tpgan} proposed a dual-path generator that concatenates a coarsely generated frontal face with local profile facial features to generate a high-quality frontal view. \cite{zhao18_pim} extended TP-GAN to jointly learn frontal face generation and discriminative feature embeddings for end-to-end face verification. \cite{li19_m2fpa, yin20_dagan} used attention guided synthesis with part masks to frontalize profile face images. More recently, \cite{di21_dalgan} proposed a contrastive learning approach for frontalization, achieving strong performance without using additional part annotations.

We base our method on general-purpose domain adaptation algorithms tailored to the task of thermal-to-visible face synthesis.

\textbf{Multi-Spectral Face Datasets.}
Several large-scale datasets exist for the task of thermal-to-visible face synthesis. The University of Notre Dame (UND) Dataset \cite{chen03_und} contains 241 unique identities with four low-resolution images per identity. Next, the Natural Visible and Infrared Expression Database (NVIE) \cite{wang10_nvie} captures subjects eliciting a wide range of facial expressions, with and without glasses. Data are synchronized manually after data collection. More recently, two volumes of the Multi-Modal Face Database (MMFD) \cite{hu16_mmfd} have been released, which provides synchronized imagery of visible, LWIR, and Polarimetric LWIR data at variable distances from the camera. ULFMT \cite{ghiass18_ulfmtv} contains unsynchronized MWIR and visible video recordings of 238 subjects capturing under variable conditions. The Tufts Face Database \cite{panetta20_tufts} is a multi-modal dataset, capturing 100 subjects using LWIR, NIR, 2D, computer sketches, and 3D point clouds. In this paper, we consider the case of paired LWIR and visible images. Following the protocol established in prior works \cite{he17_ganvfs, mallat19_cascade, di21_attributeguided}, we split the paired data in an 80-20 train-test split. The ARL-VTF dataset \cite{poster21_arlvtf} is a recently introduced large-scale multi-spectral face dataset, containing time-synchronized paired LWIR and visible face with ground truth facial landmark annotations. This dataset contains 395 subjects under variable pose and expression.

\section{Align, Translate, and Classify}
Our proposed identity preserving thermal-to-visible face synthesis method is based on three key ideas: (1) keypoint-based face alignment, (2) pixel-level correspondence, and (3) feature-level identity classification. (1) is applied before training the face synthesis model, while (2) and (3) are constraints enforced during model training. We apply these three ideas to off-the-shelf domain adaptation methods, specifically Pix2Pix \cite{isola17_pix2pix}, CycleGAN \cite{zhu17_cyclegan}, and CUT \cite{taesung20_cut}. 

\subsection{Keypoint-Based Face Alignment}
\begin{figure}[t]
    \centering
    \includegraphics[trim=1.5cm 10.7cm 8cm 4cm, clip, width=0.95\linewidth]{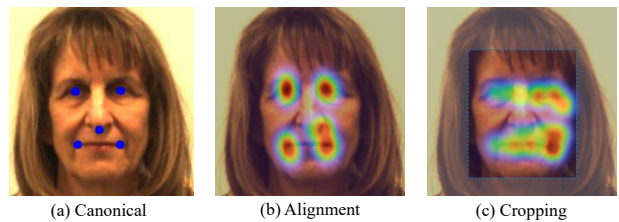}
   \vskip-11pt \caption{We qualitatively demonstrate the impact of keypoint-based face alignment by plotting the distribution of ground truth keypoints (eyes, tip of the nose, and corners of the mouth) after performing alignment and cropping, respectively, for profile-pose examples of an individual in the ARL-VTF dataset. Note that the cluster of keypoints after alignment (subfigure b) are closely correlated with the canonical keypoints for a front-pose face (subfigure a) while cropping (subfigure c) does not constrain keypoint locations. Tight clustering between samples due to alignment significantly improves both face synthesis and feature extraction.}
    \label{fig:alignment}
\end{figure}

Keypoint-based face alignment is a widely used technique to improve visible-spectrum face verification \cite{ranjan17_ultraface, bansal17_cnnbasedface, deng19_arcface}. Alignment transforms the image such that key facial landmark locations (e.g., eyes, nose, mouth) are approximately constant in all images. Importantly, face alignment helps simplify face synthesis, particularly improving performance in synthesizing profile faces by reducing inter-sample variation, as shown in Figure \ref{fig:alignment}. In addition, we find that synthesizing aligned faces further minimizes the domain shift for feature extraction as visible-spectrum verification pipelines typically align faces before training. Surprisingly, we find that current thermal-to-visible face synthesis methods do not perform face alignment.

Face alignment consists of finding a similarity transform that maps facial keypoints onto standard locations of a fixed-size image. In face recognition, this is commonly done by determining the similarity transform that best maps a collection of estimated or annotated face keypoints to predefined positions. These predefined positions are produced empirically by averaging over a subset of forward-facing samples with hand-labeled keypoints. In this work, we use ground-truth keypoints to train and evaluate performance on the ARL-VTF dataset, and estimate keypoints using a detector, as described in Section \ref{sec:detection_kp}, for the TUFTS and MILAB-VTF(B) datasets to perform face alignment. The similarity transform is solved using singular value decomposition as described in~\cite{umeyama1991least}.

\subsection{Pixel-Level Correspondence}
Paired multi-spectral datasets are often calibrated to ensure pixel-wise correspondence between domains to facilitate supervised learning. Since we are interested in using off-the-shelf domain adaptation algorithms, many of which are unsupervised, we modify these methods to explicitly constrain the generated image to minimize the $\ell_1$ distance with the ground truth visible image. Interestingly, we find that this additional regularization complements unsupervised image synthesis algorithms, as shown in Section \ref{sec:ablation}. 

\subsection{Identity Classification with Label Smoothing}
\begin{figure}[t]
    \centering
    \includegraphics[trim=0.5cm 10cm 11cm 3.5cm, clip, width=0.95\linewidth]{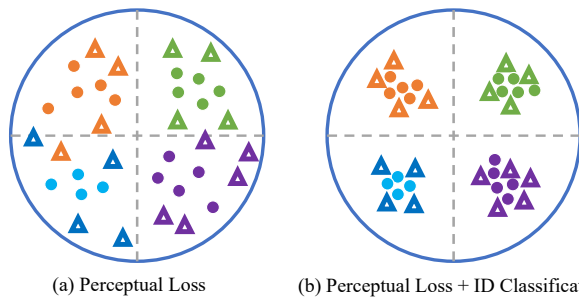}
   \vskip-13pt \caption{The identity loss described in \cite{di19_selfattn} attempts to minimize the $\ell_2$ distance between the synthesized image features and real image features. This constraint assumes that features from synthetic faces of the same identity are close together in feature space. Although this is valid for easy examples (shown as circles), it may not hold for hard examples (shown as triangles).}
    \label{fig:cluster}
\end{figure}

Reducing inter-identity distance in feature space has been shown to improve face verification from synthesized images. We extend the perceptual loss presented in \cite{di19_selfattn} by enforcing an additional constraint such that all generated images of a particular class must cluster together. We illustrate the effect of this additional constraint in Figure \ref{fig:cluster}. \cite{luo19_bagoftricks, khorramshahi20_selfsupervised} show that for datasets with a limited number of identities, as is the case in many paired thermal-to-visible datasets, label-smoothing in the cross-entropy loss effectively prevents over-fitting. We describe the identity loss function below:
\begin{equation*}
\begin{aligned}
  \mathcal{L}_C &= 1 - \cos{(\Phi_F(\Phi_G(x_t)), \Phi_F(x_v))} + \sum_{i=1}^{N} -q_i \log p_i \\
  q_i &= \begin{cases} 
      1 - \frac{N - 1}{N} \epsilon &  i = y \\
    \frac{\epsilon}{N} & o.w. \\

    \end{cases}
  \end{aligned}
\end{equation*}

where $\Phi_F$ is a pretrained visible-spectrum face feature extractor, $\Phi_G$ is the thermal-to-visible synthesis network, $x_t$ is the input thermal image, $x_v$ is the ground truth visible image, $\epsilon$ is a noise constant, and $N$ is the total number of classes. We train a 3-layer MLP $\phi_C$ to classify identity labels given features extracted from the synthesized images such that $p_i = \phi_C(\phi_F(\Phi_G(x_t)))$. In practice, the perceptual loss minimizes the cosine distance between normalized real and synthetic features. However, this constraint is ineffective if the features extracted from the ground truth image are not discriminative. The cross-entropy loss additionally ensures that all synthesized faces, particularly hard examples such as profile faces, are close in the embedding space.

\subsection{Generic Formulation}
We can generalize our proposed method as follows:
\begin{equation*}
\begin{aligned}
  \mathcal{L} &= \lambda_1 \cdot \mathcal{L}_G + \lambda_2 \cdot \mathcal{L}_1 + \lambda_3 \cdot \mathcal{L}_C
  \end{aligned}
\end{equation*}
where $\lambda_1, \lambda_2, \&  \ \lambda_3 $ are regularization constants, $\mathcal{L}_G$ is the loss function of a general-purpose domain-adaptation algorithm, $\mathcal{L}_1$ is the pixel-wise loss between the synthesized and target images, and $\mathcal{L}_C$ is the identity loss term described above. Our ablation study in Section \ref{sec:ablation} suggests that each component is necessary independent of the synthesis network used. 

\section{MILAB-VTF(B): A large-scale dataset for unconstrained thermal-to-visible synthesis}
\begin{figure}[b]
    \vspace{-1cm}
    \centering
    \includegraphics[trim=0cm 6cm 11cm 1.5cm, clip, width=0.9\linewidth]{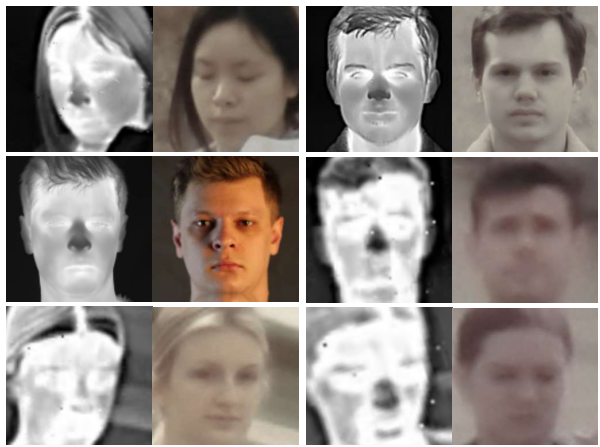}
    \vskip-5pt\caption{We present representative examples of paired thermal-visible data from the MILAB-VTF(B) dataset. Importantly, this dataset contains diverse lighting and image quality due to the large range of collection distances not found in other multi-spectral datasets.}
    \label{fig:mag400_example}
\end{figure}

In this section, we present MILAB-VTF(B), a challenging large-scale multi-spectral face dataset. This dataset captures unsynchronized paired thermal and visible data from 400 identities in varied poses and distances. Sample images from this dataset are shown in Figure~\ref{fig:mag400_example}. We compare MILAB-VTF(B) to other multi-modal datasets in Figure \ref{fig:datasets}.

{
\setlength{\tabcolsep}{0.27em} 

\begin{figure*}[t]
\centering
\begin{subfigure}
    \centering
    \includegraphics[trim=0cm 5.9cm 0cm 5.9cm, clip, width=0.85\linewidth]{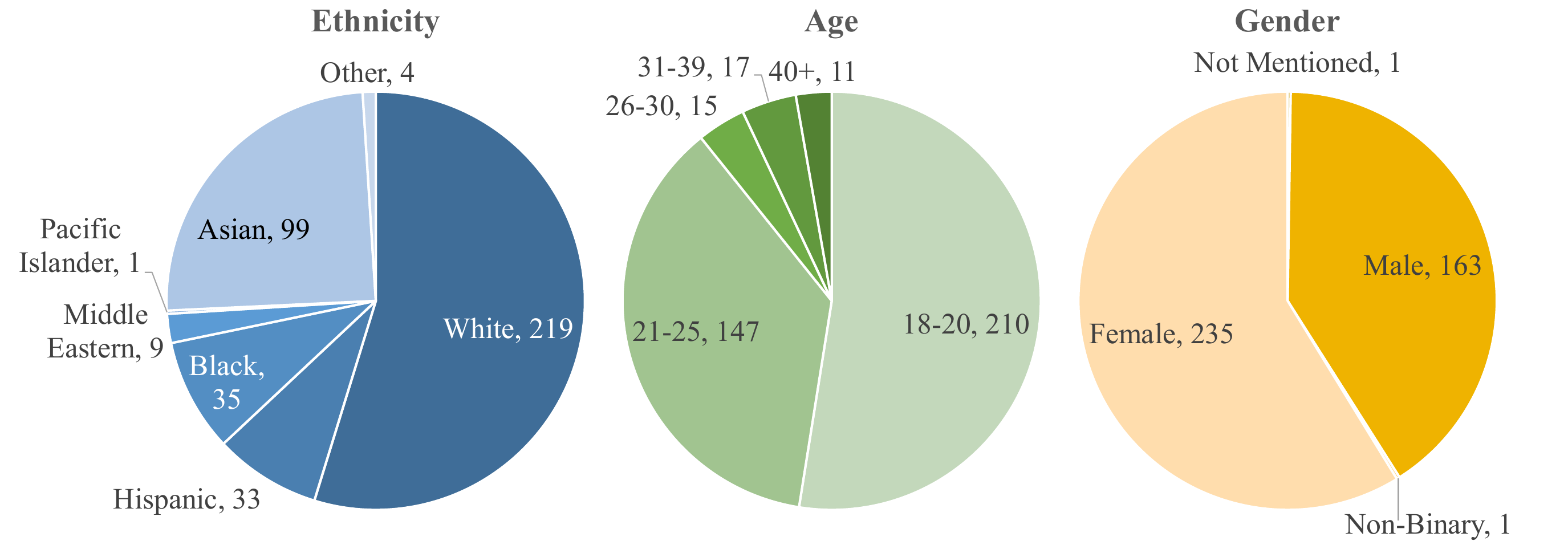}
\end{subfigure}

\begin{subtable}
\centering
\begin{tabular}{l c c c c c c}
\toprule
\multirow{1}{*}{Dataset} & \multicolumn{1}{c}{Modalities} & \multicolumn{1}{c}{Subjects} & \multicolumn{1}{c}{Variability} & \multicolumn{1}{c}{IR Resolution (W $\times$ H)} & \multicolumn{1}{c}{Range (m)}  \\ 
\toprule
UND \cite{chen03_und} & LWIR, Visible & 241 & I, E, T & 320 $\times$ 240 & Unspecified \\
NVIE \cite{wang10_nvie} & LWIR, Mono & 215 & I, E, G & 320 $\times$ 240 & 0.75 \\
ULFMT \cite{ghiass18_ulfmtv} & MWIR, Visible & 238 & P, E, T, G & 640 × 512 & 1.0 \\
ARL-MMFD \cite{hu16_mmfd} & P-L, LWIR, Visible & 111 & E & 640 $\times$ 480 (LW) & 2.5, 5.0, 7.5 \\
Tufts \cite{panetta20_tufts} & NWIR, LWIR, Visible & 100 & P, E & 336 $\times$ 256 & 1.5 \\
ARL-VTF \cite{poster21_arlvtf} & LWIR, Visible, Mono & 395 & P, E, G & 630 $\times$ 512 & 2.1 \\
\textbf{MILAB-VTF(B)} & MWIR, Visible & 400 & P, L, I$_N$, E$_N$, O$_N$ & 1280 $\times$ 1024 & 1.5, 100, 200, 300, 400 \\

\bottomrule
\end{tabular}
\end{subtable}

\caption{We denote the variable characteristics of each dataset as follows: (P)ose, (I)llumination, (E)xpression, (T)ime-lapse, (G)lasses, (O)cclusion, and (L)ocation. We use the subscript $N$ to identify characteristics that occur due to natural outdoor conditions (i.e. sunlight, clouds, and wind). MILAB-VTF(B) uniquely captures high-resolution paired thermal and visible scenes outdoors at large distances. Importantly, the dataset is diverse with respect to ethnicity, age, and gender. This table is adapted from \cite{poster21_arlvtf}. 
}
\label{fig:datasets}
\end{figure*}
}

\textbf{Dataset Collection.}
The data was collected over five weeks, starting in February 2021. The dataset contains 400 subjects, each of whom completed an Institutional Review Board (IRB) approved consent form before image acquisition. The dataset has unsynchronized paired thermal and visible videos indoors at 1.5 meters and outdoors at 100, 200, 300, and 400 meters. Each subject was recorded for approximately 30 seconds. Participants were instructed to articulate their heads $\pm 60^\circ$ left-and-right and up-and-down in each scene. Indoor scenes were captured using a Canon Mark IV (Visible) and a FLIR A8581 (MWIR). Outdoor scenes were captured by a Nikon P1000 (Visible) and a FLIR RS8513 (MWIR).

\textbf{Training and Evaluation Protocol.}
The MILAB-VTF(B) dataset provides unsynchronized, paired thermal-visible videos and anonymized identifiers for each subject. We provide algorithmically generated frame synchronization between thermal and visible videos, face bounding boxes, and keypoints, which will be useful for developing end-to-end multi-spectral face verification pipelines.

We select 320 identities for training and sequester 80 identities for evaluation. Following standard face verification protocols, we create gallery and query sets from the sequestered data. Specifically, we create four non-overlapping galleries and four non-overlapping query sets by splitting the evaluation data by pose (i.e., frontal/profile) and location (i.e. indoor/outdoor).

\section{End-to-End Thermal-to-Visible Synthesis}
In this section, we briefly describe our end-to-end thermal-to-visible face verification pipeline. First, we present the face detection and keypoint regression models required to pre-process data before thermal-to-visible synthesis. Next, we highlight our automatic temporal alignment algorithm for synchronizing the visible and thermal data streams in the MILAB-VTF(B) dataset. We leverage this end-to-end pipeline to train and evaluate both the TUFTS \cite{panetta20_tufts} and MILAB-VTF(B) datasets.

\subsection{Face Detection and Keypoint Regression}
\label{sec:detection_kp}
Face localization and keypoint regression are essential pre-processing steps in end-to-end thermal-to-visible face verification. Importantly, we find that we can jointly train a single model to address both thermal and visible domains, improving label efficiency \cite{poster21_landmarktransfer}. Given the limited availability of face bounding boxes and keypoint detections in the thermal domain, we can use models trained on visible images to bootstrap multi-spectral domain training. 
Using a large set of publicly available face bounding boxes for visible images and a small set of face bounding boxes for thermal images from the ARL-VTF and MILAB-VTF(B) datasets, we train a Faster-RCNN \cite{ren15_fasterrcnn} detector to localize faces. 
We propose a novel multi-task architecture for facial landmark localization. Our keypoint regressor outputs four sets of keypoints (55, 45, 21, 5) at different locations using a truncated ResNet-50 \cite{he16_resnet} backbone, followed by four different encoder-decoder heads, representing different densities of keypoints, and a fifth decoder that predicts yaw-roll-pitch.  We have implemented a inference engine that takes as input a predicted face bounding box, generates five random crops of the detected face, and runs the keypoint detector on each crop. From these five crops, we predict five sets of keypoints that are aggregated using RANSAC. 

\subsection{Pairwise Image Synchronization}
Since the MILAB-VTF(B) dataset does not provide temporally aligned paired videos, we propose a simple synchronization algorithm using keypoint-based face alignment. We refer the reader to \cite{umeyama1991least} for the alignment optimization.

For each frame in paired thermal and visible videos, we extract facial landmarks using our proposed keypoint regession model. We transform the keypoints into the aligned domain using a similarity transform obtained using least-squares optimization. Transforming both thermal and visible keypoints into the aligned domain implies that keypoints from corresponding visible and thermal images should match. Using this insight, we perform a greedy matching between thermal and visible frames to minimize the normalized $\ell_2$ distance between the aligned visible and thermal keypoints. This temporal alignment algorithm provides us approximate thermal-visible pairs, allowing us to leverage supervised algorithms for thermal-to-visible face synthesis.

This approach allows for many-to-one matching, whereby multiple visible frames can be mapped to the same thermal frame. Moreover, this algorithm does not preserve the temporal characteristics of the input video. Since we only consider keypoints, facial expression, hair, and ocular movement between synchronized thermal and visible frames may not be consistent. However, we find that this simple approach works well in practice.

\section{Experiments and Results}

\begin{figure*}[t]
    \centering
    \includegraphics[trim=0cm 3cm 4cm 3.5cm, clip, width=0.825\linewidth]{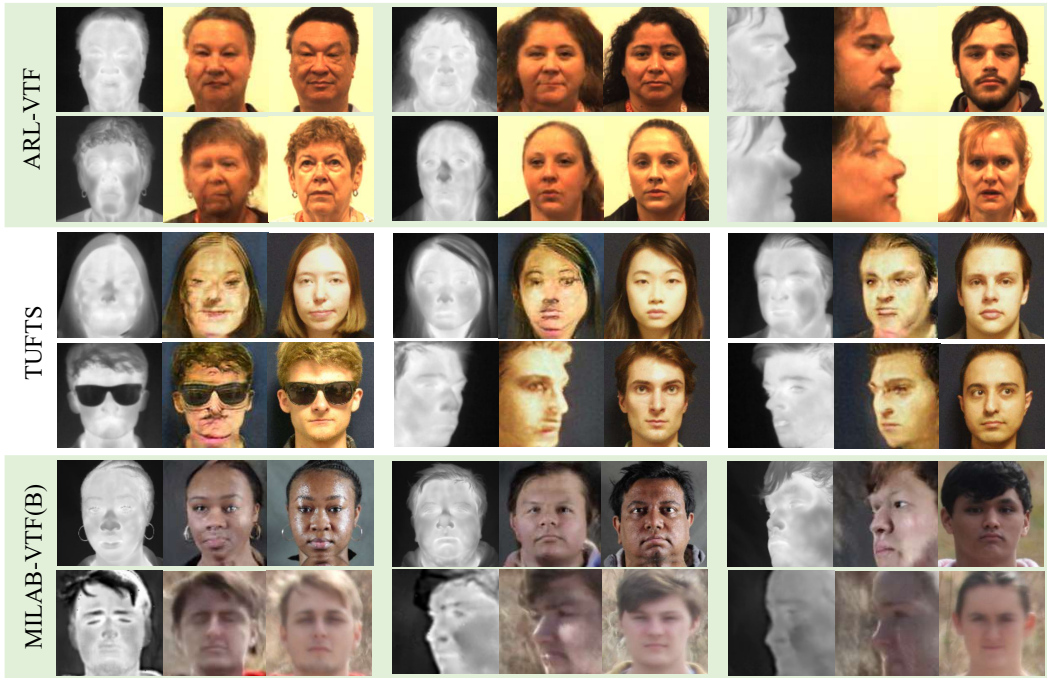}
    \vskip-10pt\caption{We share synthesized images using CUT-ATC from the ARL-VTF, TUFTS, and MILAB-VTF(B) datasets. Each three-tuple of images contains the thermal spectrum input, the synthesized output, and a frontal-pose visible image from the same identity. The limited data of the TUFTS dataset and atmospheric turbulence in the MILAB-VTF(B) dataset significantly degrade perceptual quality. Textures, particularly in hair, are not well preserved.}
    \label{fig:synthesized_example}
    
\end{figure*}
In this section, we highlight the performance of our proposed method on the ARL-VTF and TUFTS datasets and establish a strong baseline for the MILAB-VTF(B) dataset. In addition, we conduct ablation studies on the ARL-VTF dataset using three off-the-shelf domain adaptation methods \cite{isola17_pix2pix, zhu17_cyclegan, taesung20_cut}. Following standard face verification evaluation protocols, we report verification performance using the area under the curve (AUC), equal error rate (EER), and true accept rate (TAR) at false accept rates (FAR) of 1\% and 5\%. 

\textbf{Implementation Details.} 
We train each thermal-to-visible face synthesis model for twenty-five epochs and use the Adam optimizer with a learning rate of 2e-3. Importantly, we find that batch size has a significant impact on face verification performance. Smaller batch sizes work best, and we opt to train our models with a batch size of 4. Moreover, we find that training for a large number of epochs causes significant over-fitting. We synthesize 128 $\times$ 128 thumbnails and resize the output image using nearest-neighbor interpolation to match the input size requirement of the fixed feature extractor. We perform double-flip augmentations and average the extracted features before computing verification metrics. We report average performance across all gallery and query splits in the main paper and refer the reader to the supplemental material for a detailed cohort analysis.

\subsection{Ablation Study Using the ARL-VTF Dataset}
\label{sec:ablation}
We use Pix2Pix \cite{isola17_pix2pix}, CycleGAN \cite{zhu17_cyclegan}, and CUT \cite{taesung20_cut} as backbone architectures to demonstrate the broad applicability of our proposed methods, as shown in Table \ref{tab:ablation}. Our baseline models emulate current practices, simply cropping thermal faces, synthesizing visible faces, and extracting features with VGGFace \cite{parkhi15_vggface}. The batch size, learning rate, and number of training epochs remain constant through the ablation study. Each row in Table \ref{tab:ablation} builds upon the previous results, which are each described in the following sections.

{
\setlength{\tabcolsep}{0.27em} 

\begin{table}[b]
\centering
\small

\caption{Ablation Study Using the ARL-VTF Dataset}
\label{tab:ablation}
\begin{tabular}{l c c c c}

\toprule
\multirow{1}{*}{Model} & \multicolumn{1}{c}{AUC $\uparrow$} & \multicolumn{1}{c}{EER $\downarrow$} & \multicolumn{1}{c}{TAR@1\% $\uparrow$} & \multicolumn{1}{c}{TAR@5\% $\uparrow$}  \\ 
\toprule
Pix2Pix \cite{isola17_pix2pix}  & 85.3	& 22.4	& 21.6	& 43.6 \\
$\indent$ + Alignment  & 90.1	& 17.8	& 30.9	& 58.3 \\
$\indent$ + $\mathcal{L}_1$ Pixel Loss  & - & - & - & - \\
$\indent$ + Identity Loss   & \textbf{91.3} & \textbf{16}	& \textbf{33.6}	& \textbf{62.6} \\
$\indent$ w/ ArcFace  & 83.8 & 24.4	& 23 & 44.8 \\
\midrule

Cycle-GAN \cite{zhu17_cyclegan}  & 51.1	& 49 & 1.8 & 5.7 \\
$\indent$ + Alignment  & 92.1 & 14 & 54.7 & 71.9 \\
$\indent$ + $\mathcal{L}_1$ Pixel Loss  &  94	& 11.7	& 58.1	& 79.3\\
$\indent$ + Identity Loss   & 95.3	& 10.7	& 58.8	& 81.1 \\
$\indent$ w/ ArcFace & \textbf{96.8}	& \textbf{9.2}	& \textbf{68.7}	& \textbf{85.2} \\
\midrule

CUT \cite{taesung20_cut}  & 52.6 & 47.9	& 1.4 & 5.7 \\
$\indent$ + Alignment   & 93.5 & 13.6 & 52.7 & 74.6 \\
$\indent$ + $\mathcal{L}_1$ Pixel Loss  & 95.4 & 10.3 & 61.5 & 80.4\\
$\indent$ + Identity Loss  & 95.9 & 9.9 & 64.3 & 82.6 \\
$\indent$ w/ ArcFace & \textbf{97.7} & \textbf{6.9} & \textbf{77.2} & \textbf{90.5} \\

\bottomrule
\end{tabular}
\end{table}
}

\textbf{Impact of Face Alignment.}
Many standard thermal-to-visible face synthesis algorithms tightly crop a subject’s face to isolate facial features for synthesis. However, we find that aligning the keypoints to canonical locations before synthesis is an important pre-processing step that considerably impacts perceptual quality and face verification performance. We find that alignment improves performance across all models. Importantly, we highlight that alignment produces significant improvements in unsupervised models. AUC increases by more than 40\% for both CycleGAN and CUT. Surprisingly, we find that CycleGAN and CUT, both unsupervised domain adaptation algorithms perform better than Pix2Pix, indicating that further study into unsupervised algorithms may be warranted. We posit that unsupervised synthesis is easier after performing alignment because key facial features are always in approximately the same location, allowing an unsupervised model to learn more discriminative features. Moreover, unsupervised losses seem to capture identity preserving characteristics by requiring consistency between thermal and visible images. 

\textbf{Impact of Pixel-wise Correspondence.} 
Given that many thermal-to-visible synthesis datasets provide corresponding images in both domains, it is natural to extend unsupervised algorithms by enforcing an additional constraint such that the synthesized image minimizes the $\ell_1$ distance with the ground truth visible image. We extend both CycleGAN and CUT such that the generated images minimize the pixel-wise distance from the ground truth and observe a modest performance improvement. This improvement indicates that the supervisory signal from the $\ell_1$ pixel-wise loss is complementary to the unsupervised losses in both CycleGAN and CUT, respectively. Note that since Pix2Pix already enforces this constraint, we omit this row.

\textbf{Impact of Identity Loss.} 
Both prior modifications improve the visual quality of generated images but do not explicitly constrain the resulting feature embedding. Here, we minimize the cosine distance between the features from the generated visible image and the real visible image, this groups the features of the generated image close to the features of the real images. In addition, we also use cross-entropy loss with label smoothing to classify the features from the generated image into one of the $N$ classes in the training set. This additional loss minimizes the inter-class distance despite variations in pose and expression. We set $\epsilon$=0.1 and $N$=236 for this study. Importantly, we note that label smoothing is an effective way to prevent overfitting on the training set, facilitating greater generalization \cite{luo19_bagoftricks, khorramshahi20_selfsupervised}. Table \ref{tab:ablation} highlights that this feature classification provides small but consistent improvement across all tested models.

\textbf{Impact of Feature Extractor.}
Many state-of-the-art algorithms use VGGFace \cite{parkhi15_vggface} to extract discriminative features from the penultimate fully-connected layer. To facilitate a fair comparison with other methods, we perform all ablation studies using this feature extractor unless explicitly noted. Importantly, the improved performance of our baseline is not just a result of using a more robust feature extractor, but rather due to effective domain adaptation. Surprisingly, we find that ArcFace \cite{deng19_arcface}, a feature extractor that improves considerably upon VGGFace on visible domain benchmarks, performs worse when extracting features from images synthesized using Pix2Pix. This suggests that the effectiveness of the feature extractor is dependent on the network architecture. We posit that ArcFace is more sensitive to domain shift caused by synthetic imagery, indicating that CycleGAN and CUT more closely approximate the distribution of real faces. 

\subsection{State-of-the-Art Comparison}
 We compare the proposed method against general-purpose methods and task-specific methods that focus on frontal-to-profile matching. We find that our proposed method significantly improves upon the state-of-the-art. As shown in Table \ref{tab:ablation}, CUT, augmented by our proposed modifications, performs best. We compare this model, denoted as CUT-ATC, with other leading methods on the ARL-VTF and TUFTS face datasets.  

{
\setlength{\tabcolsep}{0.27em} 

\begin{table}[h]
\centering
\small


\caption{ARL-VTF State-of-the-Art Comparison}
\label{tab:sota_vtf}
\begin{tabular}{l c c c c}
\toprule
\multirow{1}{*}{Model} & \multicolumn{1}{c}{AUC $\uparrow$} & \multicolumn{1}{c}{EER $\downarrow$} & \multicolumn{1}{c}{TAR@1\% $\uparrow$} & \multicolumn{1}{c}{TAR@5\% $\uparrow$}  \\ 
\toprule
GANVFS \cite{he17_ganvfs} & 85.7 & 20.0 & 42.5 & 57.9 \\
Multi-AP-GAN \cite{di21_attributeguided} & 87.0 & 18.1 & 49.1 & 63.4 \\
Fondje et. al. \cite{fondje20_crossdomain} & 90.1 & 13.1 & 73.0 & 78.8 \\
CUT-ATC & \textbf{97.7} & \textbf{6.9} & \textbf{77.2} & \textbf{90.5} \\
\bottomrule
\end{tabular}

\end{table}
}

Our proposed model improves upon recent methods for thermal-to-visible synthesis. Table \ref{tab:sota_vtf} demonstrates that we improve on all metrics, specifically increasing the AUC by 7.6\%, reducing EER by 6.2\%, and improving TAR at FAR=1\% and FAR=5\% by 4.2\% and 11.7\%, respectively. Fondje et. al. \cite{fondje20_crossdomain} perform better than prior methods because they align their input images similar to our proposed approach. However, rather than trying to learn domain agnostic features from a small dataset, we extract features from synthesized faces using ArcFace \cite{deng19_arcface}.

{
\setlength{\tabcolsep}{0.27em} 

\begin{table}[h]
\centering
\small

\caption{TUFTS State-of-the-Art Comparison}
\label{tab:sota_tufts}
\begin{tabular}{l c c c c}
\toprule
\multirow{1}{*}{Model} & \multicolumn{1}{c}{AUC $\uparrow$} & \multicolumn{1}{c}{EER $\downarrow$} & \multicolumn{1}{c}{TAR@1\% $\uparrow$} & \multicolumn{1}{c}{TAR@5\% $\uparrow$}  \\ 
\toprule
GANVFS \cite{he17_ganvfs} & 73.8 & 32.3 & - & - \\
CRN + CL \cite{mallat19_cascade} & 74.9 & 31.7 & - & - \\
Multi-AP-GAN \cite{di21_attributeguided} & 77.4 & 29.9 & - & - \\
CUT-ATC & \textbf{87.4} & \textbf{21.3} & \textbf{26.2} & \textbf{50.6} \\

\bottomrule
\end{tabular}
\end{table}
}

The TUFTS face dataset is  challenging due to the limited number of training examples. Despite this limitation, Table \ref{tab:sota_tufts} shows that our proposed method similarly improves on all reported metrics. Specifically, we improve the AUC by 10.0\% and reduce EER by 8.5\%. With only 1200 paired visible-thermal training images, over-fitting is a significant concern for this dataset.

{
\setlength{\tabcolsep}{0.27em} 

\begin{table}[h]
\centering
\small

\caption{ARL-VTF Comparison with Face Frontalization Methods \ Gallery 0010}
\label{tab:sota_pose_vtf}
\begin{tabular}{l c c c c}
\toprule
\multirow{1}{*}{Query 00 Pose} & \multicolumn{1}{c}{AUC $\uparrow$} & \multicolumn{1}{c}{EER $\downarrow$} & \multicolumn{1}{c}{TAR@1\% $\uparrow$} & \multicolumn{1}{c}{TAR@5\% $\uparrow$}  \\ 

\toprule
PIM \cite{zhao18_pim} & 68.7 & 36.6 & 5.4 & 16.5\\
M2FPA \cite{li19_m2fpa} & 75.0 & 32.3 & 5.7 & 20.3 \\
DA-GAN \cite{yin20_dagan} & 75.6 & 31.2 & 6.9 & 22.2 \\
DAL-GAN \cite{di21_dalgan} & 77.5 & 29.1 & 8.2 & 25.9\\
CUT-ATC & \textbf{94.1} & \textbf{12.1} & \textbf{63.0} & \textbf{80.9} \\
\end{tabular}
\begin{tabular}{l c c c c}
\toprule
\multirow{1}{*}{Query 10 Pose} & \multicolumn{1}{c}{AUC $\uparrow$} & \multicolumn{1}{c}{EER $\downarrow$} & \multicolumn{1}{c}{TAR@1\% $\uparrow$} & \multicolumn{1}{c}{TAR@5\% $\uparrow$}  \\ 
\toprule
PIM \cite{zhao18_pim} & 73.3 & 32.7 & 6.0 & 20.4 \\
M2FPA \cite{li19_m2fpa} & 77.0 & 29.8 & 9.5 & 23.4 \\
DA-GAN \cite{yin20_dagan} & 75.8 & 30.7 & 8.4 & 23.6 \\
DAL-GAN \cite{di21_dalgan} & 82.2 & 25.1 & 10.8 & 30.6 \\
CUT-ATC & \textbf{95.6}	& \textbf{11.0} & \textbf{67.9}	& \textbf{82.7} \\
\bottomrule
\end{tabular}

\end{table}
}

Frontal-to-profile matching is a difficult sub-problem within face verification because there is little spatial overlap between both poses.  Moreover, thermal-to-visible synthesis adds additional complexity, requiring matching between synthesized and real faces. Face frontalization methods explicitly attempt to address this sub-problem by synthesizing frontal faces given a profile view for better spatial overlap when matching identities. In general, these task-specific methods are usually more effective at frontal-to-profile matching than general-purpose thermal-to-visible face synthesis approaches. However, our proposed modifications, particularly aligning profile faces to canonical coordinates, improves matching performance, as demonstrated in Table \ref{tab:sota_pose_vtf}. In particular, we improve the AUC by approximately 15\%, reduce EER by 15\%, and improve TAR at both FAR=1\% and FAR=5\% by approximately 50\% on the ARL-VTF dataset. Applying our general-purpose modifications to existing face frontalization methods may result in even better profile-to-frontal matching performance and warrants further investigation.

{
\setlength{\tabcolsep}{0.27em} 

\begin{table}[h]
\centering
\small

\caption{TUFTS Comparison with Face Frontalization Methods \ Gallery Frontal}
\label{tab:sota_pose_tufts}
\begin{tabular}{l c c c c}
\toprule
\multirow{1}{*}{Query Profile} & \multicolumn{1}{c}{AUC $\uparrow$} & \multicolumn{1}{c}{EER $\downarrow$} & \multicolumn{1}{c}{TAR@1\% $\uparrow$} & \multicolumn{1}{c}{TAR@5\% $\uparrow$}  \\ 
\toprule
PIM \cite{zhao18_pim} & 72.8 & 34.1 & 8.8 & 21.0 \\
M2FPA \cite{li19_m2fpa} & 75.1 & 31.2 & 8.3 & 23.4 \\
DA-GAN \cite{yin20_dagan} & 75.2 & 31.1 & 10.4 & 26.2 \\
DAL-GAN \cite{di21_dalgan} & 78.7 & 28.4 & 10.4 & 27.1 \\
CUT-ATC &\textbf{87.6} & \textbf{21.1} &\textbf{23.9} & \textbf{51.7} \\

\bottomrule
\end{tabular}

\end{table}
}

Similarly, we find that our proposed method improves over face frontalization methods on the TUFTS dataset. As shown in Table \ref{tab:sota_pose_tufts} we improve the AUC by 8.9\%, reduce EER by 7.3\%, and improve TAR at FAR=1\% and FAR=5\% by 13.5\% and 24.6\%, respectively. We highlight that our proposed approach remains effective despite the small size of the TUFTS dataset.

\subsection{Baseline on MILAB-VTF(B)}
We train our proposed models on the MILAB-VTF(B) dataset and report the average performance on each of the evaluation protocols. We denote the best-performing model and feature extractor combination from Table \ref{tab:ablation} for each backbone architecture with the ATC identifier.

{
\setlength{\tabcolsep}{0.27em} 

\begin{table}[h]
\centering
\small

\caption{MILAB-VTF(B) Dataset Baseline Performance}
\label{tab:mag400}
\begin{tabular}{l c c c c}
\toprule
\multirow{1}{*}{\textbf{Model}} & \multicolumn{1}{c}{AUC $\uparrow$} & \multicolumn{1}{c}{EER $\downarrow$} & \multicolumn{1}{c}{TAR@1\% $\uparrow$} & \multicolumn{1}{c}{TAR@5\% $\uparrow$}  \\ 
\toprule
Pix2Pix-ATC & 59.3 & 43.4 & 2.7 & 10.5 \\
CycleGAN-ATC & 54.9 & 46.4 & 1.5 & 7.1 \\
CUT-ATC & \textbf{68.8} & \textbf{36.3} & \textbf{8.0} & \textbf{22.3} \\
\bottomrule
\end{tabular}

\end{table}
}

Based on the evaluation results in Table \ref{tab:mag400}, MILAB-VTF(B) is more challenging than prior datasets. Our method, which achieves the state-of-the-art for ARL-VTF and TUFTS, performs significantly worse on MILAB-VTF(B). This is likely due to atmospheric turbulence and other visual artifacts resulting from long-distance data capture. Moreover, perfect pixel-wise correspondence is not guaranteed in MILAB-VTF(B) because we use an imperfect keypoint detector. This issue is further exacerbated for tiny faces, where a few pixels of error can cause significant misalignment. We did not specifically design our method to address the unique challenges presented by MILAB-VTF(B). However, we look forward to future investigations.

\section{Conclusion}
In this paper, we present an algorithm and a new challenging dataset for thermal-to-visible face verification. Through extensive experimentation, we show that appropriate modifications to off-the-shelf domain adaptation algorithms are widely applicable and significantly improve upon the state-of-the-art. Importantly, our experimental results suggest that face alignment, rather than cropping, is a simple modification that should be embraced in future visible-to-thermal synthesis and verification works. Despite the effectiveness of our baseline, MILAB-VTF(B) remains a challenging dataset that merits further exploration. 

\section{Acknowledgements}
This work was partially supported by an STTR Phase II contract W911QX20C0022 from the US Army Research Laboratory, Adelphi, MD.

\printbibliography

\section{Appendix}

\begin{figure*}[ht]
    \centering
    \includegraphics[trim=0cm 10cm 1cm 1cm, clip, width=0.95\linewidth]{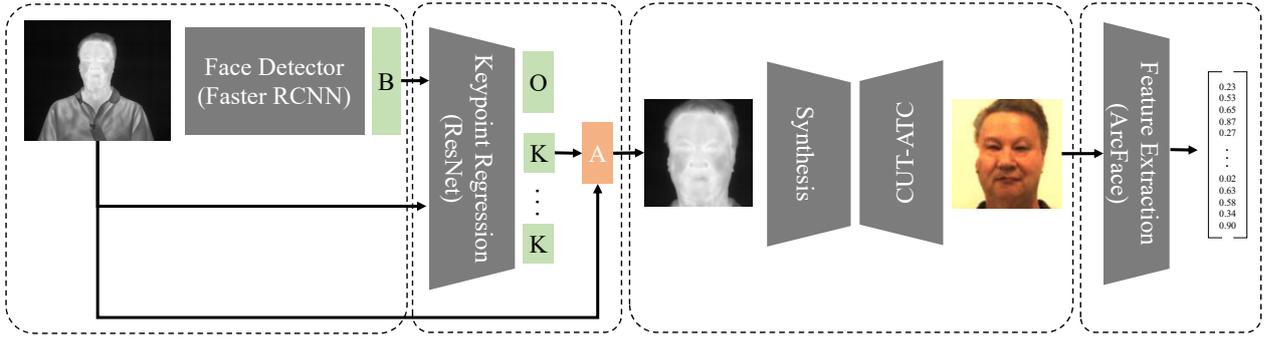}
   \vskip-10pt\caption{The thermal-to-visible synthesis pipeline contains four steps. First, we detect a bounding box around the subject's face. Next, we use the coarse-grained bounding box (Output B) to regress keypoint locations (Output K). We use these estimated landmarks to align the input image (Function A). Optionally, if the input image is from the thermal domain, we use the aligned thermal face image and synthesize a visible spectrum output. Lastly, we extract deep features that can be used to verify the subject's identity.}
    \label{fig:pipeline}
\end{figure*}

\textbf{End-to-End Inference Procedure.} Performing thermal-to-visible face verification requires a four stage pipeline: detection, keypoint regression, synthesis, and feature extraction. We demonstrate this procedure in Figure \ref{fig:pipeline} for a sample LWIR image from the ARL-VTF dataset. 

\begin{figure}[h]
    \centering
    \includegraphics[trim=1cm 7cm 11cm 3cm, clip, width=\linewidth]{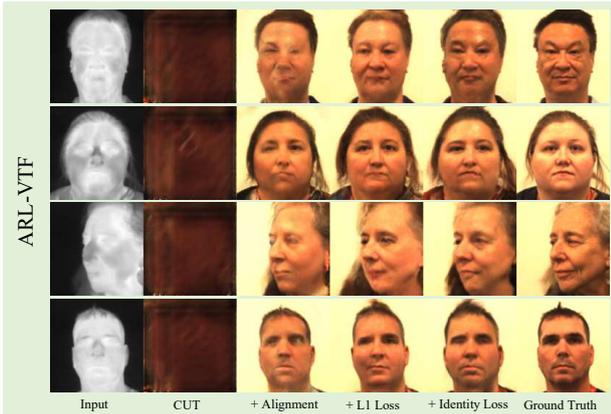}
    \vskip-10pt\caption{We note that directly applying CUT for thermal-to-visible synthesis is ineffective due to the large input image pose variation. Face alignment is essential to generate reasonable quality faces. Both $L_1$ and Identity loss functions further refine the generated image. Given the limited information present in the thermal image, we expect that perfect visible image reconstruction is unreasonable.}
    \label{fig:ablation}
\end{figure}

\textbf{Qualitative Results from Ablation Study.} In Figure \ref{fig:ablation}, we examine the qualitative effect of the three proposed modifications to CUT. Training hyperparameters are the same for all experiments. First, we note that directly using CUT for thermal-to-visible face synthesis is ineffective. The GAN is unable to produce face-like images due to the large pose variation. Next, we see that aligning the input image dramatically improves synthesis quality. Lastly, we note that the $L_1$ and identity loss removes visual artifacts around discriminative facial features such as the eyes and mouth.

\begin{figure}[h]
    \centering
    \includegraphics[trim=0cm 9cm 11cm 3cm, clip, width=0.95\linewidth]{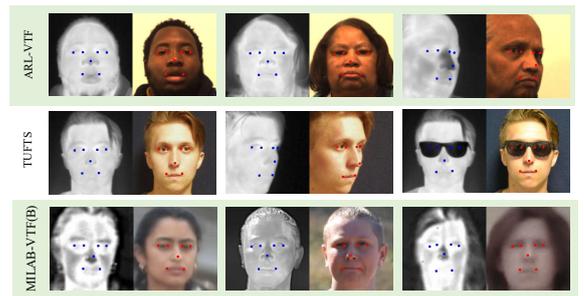}
    \vskip-10pt\caption{We show a subset of the keypoint regression estimates used for face alignment. Specifically, the keypoint regressor is trained to localize both corners of the eyes, the tip of the nose, and  corners of the mouth for both thermal and visible face images. We use a single model for both thermal and visible images.}
    \label{fig:keypoint}
\end{figure}

\textbf{Evaluating Keypoint Regression}. Our qualitative and quantitative results highlight the importance of face alignment, which requires robust landmark localization. Since the ARL-VTF dataset provides ground truth annotations for five landmarks, we evaluate our keypoint regressor on the test set and compare our performance against DAN \cite{kowalski17_dan}. First, we qualitatively demonstrate the accuracy of our keypoint regression in Figure \ref{fig:keypoint}. We can accurately identify face landmarks on low-resolution faces, profile faces, and faces with occlusions (i.e. sunglasses). 

The ground truth keypoints provided by the ARL-VTF dataset annotate the center of both eyes, the base of the nose, and corners of the mouth. Since the keypoints provided by our keypoint method (as shown in Figure \ref{fig:keypoint}) do not exactly correspond to the ground truth keypoints, we estimate the center of the eyes by taking the average position of the corners of each eye. Next, we learn a linear regression model using the validation set of the ARL-VTF dataset to find an offset that best matches our predicted keypoints to the ground truth. We evaluate our model in Table \ref{tab:keypoints} and show that we are able achieve state-of-the-art performance, improving upon DAN \cite{kowalski17_dan} in all metrics.

\textbf{Cohort Analysis.} We report the breakdown analysis of CUT-ATC on the ARL-VTF (Table \ref{tab:arl_vtf_cohort}), TUFTS (Table \ref{tab:tufts_cohort}), and VTF-400 (Table \ref{tab:mag400_cohort}) datasets. We note that synthesis and verification of profile faces is the most challenging sub-task. Profile face verification is inherently challenging because of the extreme pose variation. Synthesizing profile faces for verification adds additional challenge as the imbalanced distribution of frontal faces to profile faces means that models likely generalize poorly to profile faces. Lastly, we highlight that synthesis and verification performance for outdoor scenes is lower than indoor scenes, indicating that this is a more challenging setup that necessitates further exploration.

{
\setlength{\tabcolsep}{0.27em} 
\begin{table*}[h]
\vspace{1cm}
\centering
\small

\caption{Keypoint Evaluation on ARL-VTF Dataset}
\label{tab:keypoints}
\begin{tabular}{l l c c c c c c c}
\toprule
\multirow{1}{*}{Sequence} & \multirow{1}{*}{Method} & \multicolumn{1}{c}{Mean $\downarrow$} & \multicolumn{1}{c}{STD $\downarrow$} & \multicolumn{1}{c}{Median $\downarrow$} & \multicolumn{1}{c}{MAD $\downarrow$}  & \multicolumn{1}{c}{Max Error $\downarrow$}  & \multicolumn{1}{c}{AUC$_{0.08} \uparrow$} & \multicolumn{1}{c}{Failure Rate$_{0.08} \downarrow$}\\ 

\toprule
Baseline & DAN \cite{kowalski17_dan} & 0.0326 & 0.0155 & 0.0283 & 0.0119 & 0.0857 & 0.5798 & 0.0080 \\
         & Ours & \bf{0.0192} & \bf{0.0069} & \bf{0.0179} & \bf{0.0037} & \bf{0.0631} & \bf{0.7606} & \bf{0.0}\\
\midrule
Expression & DAN \cite{kowalski17_dan} & 0.0324 & 0.0157 & 0.0276 & 0.0122 & 0.1109 & 0.5946 & 0.0076 \\
           & Ours & \bf{0.0212} & \bf{0.0095} & \bf{0.0190} & \bf{0.0047} &  \bf{0.1106} & \bf{0.7345} & \bf{0.0010} \\
\midrule
Pose & DAN \cite{kowalski17_dan} & 0.1012 & 0.0562 & 0.0949 & 0.0472 & 0.4431 & 0.1692 & 0.5868 \\
     & Ours & \bf{0.0316} & \bf{0.0186} & \bf{0.0265} & \bf{0.0086} & \bf{0.2145} & \bf{0.6116} & \bf{0.0290} \\
\midrule

\end{tabular}
\end{table*}
\vspace{\fill}
}

\clearpage
{
\setlength{\tabcolsep}{0.27em} 

\begin{table*}[h]
\centering
\small

\caption{CUT-ATC Cohort Analysis on ARL-VTF Dataset}
\label{tab:arl_vtf_cohort}
\begin{tabular}{l l c c c c}
\toprule
\multirow{1}{*}{Gallery} & \multirow{1}{*}{Query} & \multicolumn{1}{c}{AUC $\uparrow$} & \multicolumn{1}{c}{EER $\downarrow$} & \multicolumn{1}{c}{TAR@1\% $\uparrow$} & \multicolumn{1}{c}{TAR@5\% $\uparrow$}  \\ 

\toprule
0010 & 00 Baseline & 99.5 &	2.8 &	91.0 &	97.9 \\
     & 10 Baseline & 99.7 &	2.3 &	94.3 &	100.0 \\
     & 11 Baseline & 97.1 &	9.6 &	59.7 &	81.7 \\
     & 00 Expression & 99.0 &	5.2 &	84.2 &	94.8 \\
     & 10 Expression & 99.5 &	3.6 &	88.0 &	97.8 \\
     & 00 Pose & 94.1 &	12.1 &	63.0 &	80.9 \\
     & 10 Pose & 95.6 &	11.0 &	67.9 &	82.7 \\
\midrule
0011 & 00 Baseline & 99.6&	2.9&	91.4&	97.9 \\
     & 10 Baseline & 99.4&	3.4&	83.7&	97.7\\
     & 11 Baseline & 98.2&	7.4&	66.3&	86.7 \\
     & 00 Expression & 99.0&	5.2&	84.4&	94.8 \\
     & 10 Expression & 98.7&	6.2&	83.0&	93.3 \\
     & 00 Pose & 94.2&	12.0 &	63.2&	81.0 \\
     & 10 Pose & 94.2&	13.1&	61.2&	79.5 \\
\midrule
Average &  & 97.7&	6.9&	77.2&	90.5 \\
\end{tabular}
\end{table*}
}

{
\setlength{\tabcolsep}{0.27em} 

\begin{table*}[h]
\centering
\small

\caption{CUT-ATC Cohort Analysis on TUFTS Dataset}
\label{tab:tufts_cohort}
\begin{tabular}{l l c c c c}
\toprule
\multirow{1}{*}{Gallery} & \multirow{1}{*}{Query} & \multicolumn{1}{c}{AUC $\uparrow$} & \multicolumn{1}{c}{EER $\downarrow$} & \multicolumn{1}{c}{TAR@1\% $\uparrow$} & \multicolumn{1}{c}{TAR@5\% $\uparrow$}  \\ 

\toprule
Frontal & Frontal & 88.2 &	20.6 &	31.0 &	52.3 \\
        & Profile & 87.6 &	21.1 &	23.9 &	51.7 \\
\midrule
Profile & Frontal & 85.4 &	23.4 &	21.4 &	42.5 \\
        & Profile & 88.2 &	20.0 &	28.4 &	55.8 \\
\midrule
Average &         & 87.4 &	21.3 &	26.2 &	50.6 \\
\end{tabular}
\end{table*}
}

{
\setlength{\tabcolsep}{0.27em} 

\begin{table*}[t]
\centering
\small

\caption{CUT-ATC Cohort Analysis on VTF-400 Dataset}
\label{tab:mag400_cohort}
\begin{tabular}{l l c c c c}
\toprule
\multirow{1}{*}{Gallery} & \multirow{1}{*}{Query} & \multicolumn{1}{c}{AUC $\uparrow$} & \multicolumn{1}{c}{EER $\downarrow$} & \multicolumn{1}{c}{TAR@1\% $\uparrow$} & \multicolumn{1}{c}{TAR@5\% $\uparrow$}  \\ 

\toprule
Indoor Frontal & Indoor Frontal & 76.3	& 30.6	& 17.1	& 34.1
 \\
               & Indoor Profile & 68.3	& 36.8	& 9.6	& 22.4
 \\
               & Outdoor Frontal & 72.8	& 33.2	& 12.7	& 29.0
 \\
               & Outdoor Profile & 64.7	& 40.1	& 8.3	& 20.4
\\
\midrule
Indoor Profile & Indoor Frontal & 68.1	& 37	& 9.9	& 23.3
 \\
               & Indoor Profile & 66.9	& 37.2	& 4.3	& 17.8
 \\
               & Outdoor Frontal & 64.7	& 39.5	& 5.8	& 17.4
 \\
               & Outdoor Profile & 63.4	& 40.7	& 5.5	& 17.3
 \\
\midrule
Outdoor Frontal & Indoor Frontal & 74.4	& 32.2	& 12.4	& 29.1
 \\
                & Indoor Profile & 67.9	& 37.4	& 7.4	& 20.4
 \\
                & Outdoor Frontal & 75.5	& 31.1	& 12.7	& 31.3
 \\
                & Outdoor Profile & 65.8	& 38.9	& 4.8	& 18.2
 \\
\midrule
Outdoor Profile & Indoor Frontal & 68.8	& 36.4	& 6.8	& 20.6

 \\
                & Indoor Profile & 68.8	& 36.2	& 4.9	& 20.3

 \\
                & Outdoor Frontal & 69.3	& 35.8	& 4.0	& 19.4

 \\
                & Outdoor Profile & 65.8	& 38.4	& 1.7	& 15.4

 \\
\midrule
Average         &                  & 68.8	& 36.3	& 8.0	& 22.3

 \\
\end{tabular}
\end{table*}
}

\end{document}